\documentclass[twocolumn, switch]{article}

\usepackage{arxiv}

\usepackage[utf8]{inputenc} 
\usepackage[T1]{fontenc}    
\usepackage{hyperref}       
\usepackage{url}            
\usepackage{booktabs}       
\usepackage{amsfonts}       
\usepackage{nicefrac}       
\usepackage{microtype}      
\usepackage{lipsum}
\usepackage[binary-units=true]{siunitx}
\usepackage{comment}
\usepackage{stfloats}
\usepackage{graphicx}
\usepackage{multirow}
\usepackage{makecell}
\usepackage{epsfig}
\usepackage{pifont}
\usepackage{booktabs} 
\usepackage{footnote}
\usepackage[english]{babel}
\usepackage[linesnumbered,algoruled]{algorithm2e}
\usepackage{amsmath}
\usepackage{bm}
\usepackage{multicol}
\usepackage{amssymb}
\usepackage{setspace}
\usepackage{tikz}
\usepackage{subcaption}

\title{Adaptive Test-Time Augmentation for Low-Power CPU\vspace*{-0mm}}

\date{\vspace{-5ex}}

\author{
Luca Mocerino, Roberto G. Rizzo, Valentino Peluso, Andrea Calimera, Enrico Macii\\
Politecnico di Torino, 10129 Torino, Italy\vspace*{0mm}
}

\begin{document}
\twocolumn[ 
  \begin{@twocolumnfalse} 

\maketitle

\vspace{-1cm}

\begin{abstract}
Convolutional Neural Networks (ConvNets) are trained offline using the few available data and may therefore suffer from substantial accuracy loss when ported on the field, where unseen input patterns received under unpredictable external conditions can mislead the model.
Test-Time Augmentation (TTA) techniques aim to alleviate such common side effect at inference-time,
first running multiple feed-forward passes on a set of altered versions of the same input sample, and then computing the main outcome through a consensus of the aggregated predictions. Unfortunately, the implementation of TTA on embedded CPUs introduces latency penalties that limit its adoption on edge applications. To tackle this issue, we propose {\em AdapTTA}, an adaptive implementation of TTA that controls the number of feed-forward passes dynamically, depending on the complexity of the input.
Experimental results on state-of-the-art ConvNets for image classification deployed on a commercial ARM Cortex-A CPU demonstrate AdapTTA reaches remarkable latency savings, from 1.49$\times$ to 2.21$\times$, and hence a higher frame rate compared to static TTA, still preserving the same accuracy gain.

\end{abstract}
\vspace{0.1cm}

\keywords{Deep Learning \and Convolutional Neural Networks \and Test-Time Augmentation \and Embedded Systems}
\vspace{0.35cm}

\vspace{0.35cm}

\end{@twocolumnfalse}]

\section{Introduction \& Motivations}\label{sec:intro}
Learning pattern recognition models with good generalization capability is an extremely challenging task, as the training data often represents a tiny fraction of all the possible patterns. This is a main source of concern in high-dimensional problems, like those in computer vision, e.g., image classification, where covering the large variability across different samples gets unfeasible. In this regard, the advancements in deep learning, Convolutional Neural Networks (ConvNets) in particular, enabled unprecedented results~\cite{tan2019efficientnet}. Nonetheless, state-of-the-art ConvNets still suffer from accuracy drop when ported in real-life scenarios and operated on input patterns that differ substantially from those used at training time.
For instance, the most common sources of misprediction are the discrepancy in size and orientation of the objects caught in the image~\cite{hendrycks2019benchmarking}, as well as different light conditions or contrast.

Several techniques proposed in the recent literature aim to improve model generalization operating both at training time~\cite{cubuk2019autoaugment, cubuk2020randaugment} and inference time~\cite{sun2019test, kim2020model}. Among them, Test-Time Augmentation (TTA) is a valuable option for ConvNets hosted in the cloud and operated for visual tasks like image classification~\cite{krizhevsky2012imagenet, howard2013some, szegedy2015going}. It is a simple yet efficient strategy that involves the aggregation of partial predictions over a set of transformed versions of the same input image. When implemented on high-performance architecture, the cost of multiple feed-forward passes is compensated through input batching, that is, the augmented images get processed in parallel with negligible overhead (see Table~\ref{tab:motivational}). 
The same does not hold on the edge, where ConvNets are made run on mobile devices powered by low-power CPUs with limited resources~\cite{peluso2019inference,peluso2019performance}. 
Table~\ref{tab:motivational} collects a quantitative comparison, showing that batch inference raises a prohibitive latency overhead on embedded CPUs, which in turn prevents the portability of TTA. 
Indeed, batch inference gets 5.5$\times$ (batch size=5) and 11.2$\times$ (batch size=10) slower than a single inference (batch size=1), 
which is even less efficient than sequential processing. 

\begin{table}[!t]
\small
\centering
\caption{Inference latency (\si{\milli\second}) of state-of-the-art ConvNets measured at different batch sizes (1, 5, and 10) on a cloud GPU (NVIDIA Titan Xp with 3840 CUDA cores) and an embedded CPU (ARM Cortex-A53 with 4 cores).}\label{tab:motivational}
\resizebox{1\columnwidth}{!}{
\begin{tabular}{cccc|ccc}
\toprule
\multirow{3}{*}{\bf ConvNet} & \multicolumn{3}{c|}{\bf NVIDIA Titan Xp}  &\multicolumn{3}{c}{\bf ARM Cortex-A53}\\ 
& 1& 5& 10 &1 &5 &10 \\
\midrule
MobileNetV1  &18.2& 18.6 & 18.7 &53.1 & 290.6 & 569.9  \\
MobileNetV2  &12.1 &12.4 &12.9 & 44.2 & 261.8 & 513.5  \\
\bottomrule
\end{tabular}}
\vspace{-0.5cm}
\end{table}

Starting from these observations, this paper introduces {\em AdapTTA}, an adaptive implementation of TTA suited for embedded systems. Unlike static TTA strategies, where the number of modified samples fed to the ConvNet is fixed, 
AdapTTA self-regulates the number of transformations and feed-forward passes dynamically. The transformed images are generated and processed sequentially till the model achieves good confidence on the main outcome. Specifically, AdapTTA relies on the fact that different inputs come with different intrinsic complexity and the minimum number of transformations needed to reach an accurate classification changes on a sample basis. This suggests the number of feed-forward passes can be adjusted at run-time depending on the confidence level accumulated. In such a way, the processing gets faster for "easy" images, slower for the most "complex" ones. Leveraging the statistics of the input patterns, AdapTTA provides a substantial average speed-up compared to the original static approach. 
\begin{figure*}[ht]
\centering
\includegraphics[scale=0.55]{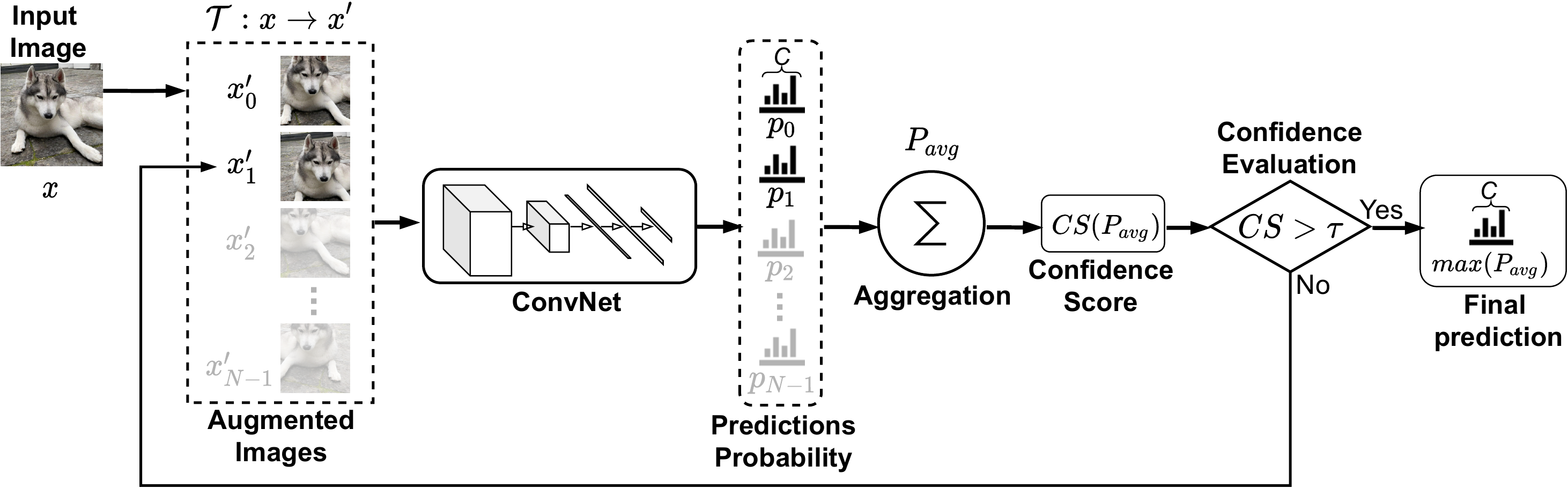}
\caption{Example of the execution flow of AdapTTA. Augmented images are generated and fed sequentially to the ConvNet. After each iteration, the predictions are aggregated, and the confidence score is computed. Depending on its value, the following transformation is applied and evaluated, or the loop is interrupted. In the example, only $x'_0$ and $x'_1$ get processed by the ConvNet to compute the final prediction. The label with the highest probability score in ${P}$\textsubscript{avg} (average between $p_0$ and $p_1$) is assigned to the input.}
\label{fig:adaptta}
\vspace{-0.5cm}
\end{figure*}
 
AdapTTA was tested on four state-of-the-art ConvNets for image classification, taking into account two common TTA policies, namely, {\em 5-Crops} and {\em 10-Crops}~\cite{krizhevsky2012imagenet, howard2013some, szegedy2015going}, which refer to five and ten consecutive crops on the same image, respectively. To notice that the main objective here is not to find the transformations that reach the highest results, a research problem already addressed in previous works~\cite{molchanov2020greedy, kim2020learning}, but rather to demonstrate the feasibility of the proposed dynamic scheme for low-power applications. AdapTTA is orthogonal to the kind of input augmentation applied indeed.
The experiments were thereby conducted on a commercial off-the-shelf embedded platform powered by an ARM Cortex-A53 CPU. Collected results show that AdapTTA reaches faster processing than static TTA, from 1.49$\times$ to 2.21$\times$ on average, still preserving the same accuracy gain. This demonstrates the improved portability and scalability of the method, which can be easily adapted to many edge applications without incurring any modification on the training pipeline.

\section{Related Works}\label{sec:related}

Data augmentation is key for training ConvNets. 
It consists of applying random transformations on the input data to increase the diversity of the training samples, with the final goal of improving the generalization capability of the model. The most simple implementations used in computer vision problems rely on a set of geometric transformation (e.g., translation, rotation, flipping) and graphical transformations (e.g., brightness, contrast, saturation), often hand-tuned by domain experts to match the conditions of real-life scenarios~\cite{krizhevsky2012imagenet, simard2003best}.
More advanced strategies aim to automate the design of the augmentation policy, for instance, through a grid search exploration~\cite{cubuk2020randaugment}, or using faster searching processes driven by reinforcement learning~\cite{cubuk2019autoaugment} and gradient-based methods~\cite{hataya2020faster}. Some of these strategies have been successfully integrated with the training of state-of-the-art ConvNets~\cite{tan2019efficientnet}.

Data augmentation at training time is often not enough to handle the unpredictable changes in the data distribution \cite{molchanov2020greedy, recht2019imagenet}. 
Therefore, TTA has been employed to increase the predictive performance of the model. 
TTA works at inference time, employing the transformations typically used in data augmentation. It aims to generate altered versions of the same input with a similar distribution of the training data, providing the model with more information. In practice, a set of modified samples is fed to the ConvNet, and the partial predictions are aggregated through majority voting or averaging. Similarly to data augmentation, the TTA policy can be hand-crafted~\cite{krizhevsky2012imagenet, howard2013some, szegedy2015going} or discovered by automatic algorithms~\cite{molchanov2020greedy, kim2020learning}. Overall, TTA enables to improve the prediction accuracy~\cite{wang2018automatic} and the robustness against adversarial attacks~\cite{volpi2018generalizing}.

Regardless of the transformations adopted, the existing TTA policies have been conceived and validated on ConvNets running in the cloud, where even a very large set of input transformations can be efficiently distributed over the extensive parallelism of GPUs.
The implementation of TTA on embedded platforms cannot leverage such parallelism, and the efficiency on low-power devices is a less explored problem, which is the target of this work.

\section{Adaptive Test-Time Augmentation}\label{sec:opt}
All the existing TTA policies share the following limitation: they apply a fixed and predefined number of transformations to each input data, namely, they are static. This represents a major bottleneck for the adoption of TTA on embedded systems.
The execution flow of a generic TTA policy for an image classification problem involving $C$ classes can be described as follows. First, a set of $N$ altered versions $x'$ of the input image $x$ is generated through the application of a set of transformation $\mathcal{T}: x \to x'$. Second, the generated images are processed by the ConvNet in parallel or sequence (more details in Section \ref{sec:results}). Third, the $N$ partial predictions are aggregated to compute the final outcome. The parameter $N$ is defined at design time by the TTA policy, hence each prediction encompasses the same number of inferences for each input image.

Such a static strategy might be too conservative for most of the samples, especially for certain inputs with key features well exposed and easy to be detected. AdapTTA has been conceived to fulfill a simple objective: implement a more flexible TTA mechanism monitoring intermediate results in order to apply the lowest number of transformations in $\mathcal{T}$ that allows to infer the correct output.

The schematic flow of AdapTTA is illustrated in Figure \ref{fig:adaptta}. Given an input image, modified versions are generated and fed to ConvNet iteratively. After each inference, the partial prediction probabilities $p_i$ are aggregated through a class-wise average ${P}$\textsubscript{avg} to compute a confidence score defined as follows: 
\begin{align}
    CS = P\textsubscript{avg-1} - P\textsubscript{avg-2}
\end{align}
where the $P$\textsubscript{avg-1} and $P$\textsubscript{avg-2} denote the probability of the first and second highest scored classes respectively. If the confidence score satisfies a pre-defined confidence threshold $\tau$ ($CS > \tau$), the prediction is deemed reliable and the TTA loop stops. The final inference outcome is then returned by taking the highest probability in ${P}$\textsubscript{avg}. The full set of augmented samples in $\mathcal{T}$ are evaluated only in the worst-case, i.e., if $CS$ gets smaller than $\tau$. In this case, AdapTTA returns the same prediction of the static TTA consuming the same computational effort.
In other words, AdapTTA implements an adaptive mechanism to adjust the augmentation passes at run-time depending on the confidence level accumulated.

Even though AdapTTA relies on a heuristic method with no guarantee of optimality in identifying the smallest subset of transformations, our experiments validate its efficiency (see Section \ref{sec:results}). The main source of non-ideality is the metric adopted to evaluate the prediction confidence, i.e., the confidence score, which we borrowed from other optimization techniques operating at run-time \cite{park2015big,tann2016runtime,mocerino2020fast}. Obviously, different aggregation functions do exists, such as majority voting, stacking \cite{breiman1996stacked}, or Bayesian methods \cite{monteith2011turning}. However, an average policy is more suited for resource-constrained devices as it requires negligible computational overhead. Further extensions of the proposed AdapTTA technique may integrate any transformation and aggregation function.

\section{Experimental Setup}\label{sec:setup}
This section describes the hardware platform used in the experiments along with the software environment adopted for the deployment. Moreover, we introduce the ConvNets taken as benchmarks and the TTA policies adopted in our comparative analysis.

\subsection{Hardware Platform and Software Configurations}\label{sec:hw}
The hardware testbench is the Odroid-C2 platform powered with the Amlogic S905 SoC. The CPU is a quad-core ARM Cortex-A53 running @1.5GHz nominal frequency. The board runs Ubuntu Mate 18.04, kernel version 3.16.72-46, released by Hardkernel.
The inference engine is TensorFlow Lite 1.14; it offers a collection of neural-network routines optimized to run on the ARM Cortex-A architecture. 
In our setup, TensorFlow Lite is cross-compiled using the GNU ARM Embedded Toolchain (version 6.5) \cite{linaro}.

\subsection{ConvNet Benchmarks}\label{sec:setup_convnets}
We adopted pre-trained ConvNets available on TensorFlow Hosted Models~\cite{tflitemodels} repository. Specifically, we considered a representative family of models for the mobile segment: {\em MobileNet}~\cite{howard2017mobilenets, sandler2018mobilenetv2}.
The networks are trained on the ImageNet dataset~\cite{deng2009imagenet} and quantized to 8-bit, a common choice for edge inference as it ensures lower memory footprint and faster processing speed with negligible accuracy loss with respect to floating-point. 
For MobileNet family, we investigated two different versions of ConvNets listed in Table~\ref{tab:benchmarks}.
The column {\bf Storage} collects the size of the \texttt{tflite}, which contains the data structures needed to deploy the model on-chip, i.e., the network weights and the topology description. The column {\bf Top-1} refers to the top-1 classification accuracy measured on the ImageNet validation set without TTA, i.e., with a standard pre-processing consisting of resizing the images to 256$\times$256 pixels and extracting the central crop of size 224$\times$224. The column $\bm{L_\textrm{\textbf{nom}}}$ reports the nominal latency of a single inference at maximum performance (4-threads @1.5GHz).

The adopted benchmarks are state-of-the-art for image classification on embedded systems. 
MobileNets are compact hand-crafted networks optimized for high performance and low memory. 

\begin{table}[!t]
\centering
\caption{Memory, nominal top-1 accuracy without TTA (Top-1), and nominal latency ($L_\textrm{\textbf{nom}}$) of the selected benchmarks.}\label{tab:benchmarks}
\begin{tabular}{lcccc}
\toprule
{\bf ConvNet} & {\bf Storage} & {\bf Top-1} & $\bm{L_\textrm{\textbf{nom}}}$ \\ 
{} & {\bf {[}MB{]}} & {\bf [\%]} & {\bf {[}ms{]}} \\
\midrule
MobileNetV1 & 4.3    &  70.0   & $53.1$   \\
MobileNetV2  &3.4    &  70.8   & $44.2$   \\
\bottomrule
\end{tabular}
\vspace{-0.5cm}
\end{table}

\subsection{Augmentation Policies}\label{sec:setup_policy}
We adopted two popular TTA policies~\cite{krizhevsky2012imagenet, howard2013some, szegedy2015going}, which are described as follows:\\
{\bf 5-Crops (5C)} - starting from a $256\times256$ image, five crops of size $224\times224$ are extracted (the center crop and the four corner crops (top-left, top-right, bottom-left and bottom-right).\\
{\bf 10-Crops (10C)} - is an extension of the 5C policy; it applies the left-to-right horizontal flipping to the five crops of 5C for a total of 10 images.

These policies enable substantial improvements in the classification accuracy as reported  Table~\ref{tab:tta_accuracy}. Obviously, doubling the number of transformations (from 5C to 10C) increases the accuracy gain.  Moreover, the transformations get processed efficiently even on resource-constrained devices: on the Cortex-A53 CPU, cropping requires only \SI{0.8}{\milli\second} and horizontal flipping \SI{0.9}{\milli\second}.

\begin{table}[!h]
    \centering
    \caption{Accuracy gain (in \%) of 5-Crops (5C) and 10-Crops (10C) TTA policies compared to the nominal values for static TTA and AdapTTA ($\tau=0.8)$.}\label{tab:tta_accuracy}
    \begin{tabular}{lcc|cc}
    \toprule
    \multirow{3}{*}{\bf ConvNet} & \multicolumn{2}{c|}{\bf Static TTA} & \multicolumn{2}{c}{\bf AdapTTA}\\
    {\bf ConvNet} & {\bf 5C} & {\bf 10C} & {\bf 5C} & {\bf 10C}\\
    \midrule
    MobileNetV1  &2.7 &3.1 &2.7 &3.1\\
    MobileNetV2  &2.2 &2.9 &2.2 &2.9\\
    \bottomrule
    \end{tabular}
\vspace{-0.5cm}
\end{table}

\section{Results}\label{sec:results}

\begin{figure*}[!t]
    \centering
    \hspace{-1cm}
    \subfloat[\label{fig:lat_mbv1} MobileNetV1]
    {\includegraphics[width=0.7\columnwidth]{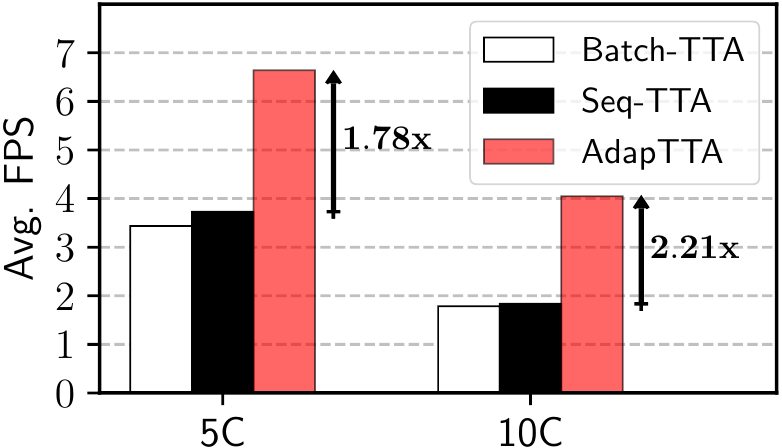}}\hspace{1.25cm}
    \subfloat[\label{fig:lat_mbv2} MobileNetV2]
    {\includegraphics[width=0.7\columnwidth]{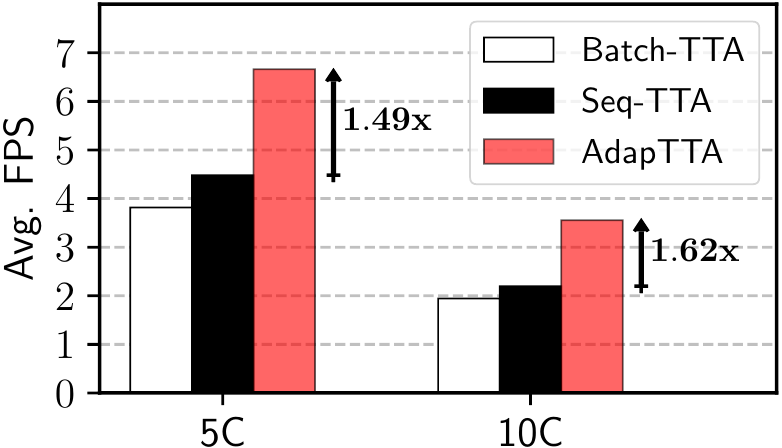}}
    \caption{Average prediction rate (Avg. FPS, higher is better) for 5C and 10C policies of the static implementations (Batch-TTA and Seq-TTA) and AdapTTA. The reported data takes into account the execution time needed for both the transformations and the inference. The arrows indicate the relative speep-up of AdapTTA compared to Seq-TTA.}
    \label{fig:accuracy_latency}
    \vspace{-5mm}
\end{figure*}

To evaluate the benefits of AdapTTA, we fixed the confidence threshold $\tau=0.8$ (maximum value is 1.0) for all the adopted ConvNets. 
Table \ref{tab:tta_accuracy} reports the accuracy gains achieved with this configuration. 
In all the benchmarks, AdapTTA ensures the same accuracy levels as static TTA. 
We point out that the selected value of $\tau$ could limit the potential savings of AdapTTA, as even slower values could be enough to achieve the same accuracy. However, we opted for this conservative choice to assess the feasibility of AdapTTA decoupling our analysis from the optimization of $\tau$.

We compared the computational efficiency of a standard static TTA and AdapTTA measuring the average prediction rate (in FPS) across the ImageNet validation set (50k images). For the static TTA, we benchmarked two different implementations:\\
\textbf{Batch-TTA} - the augmented images get processed in parallel through batching (batch size is equal to the number of crops);\\
\textbf{Seq-TTA} - the augmented images get processed sequentially. 

Collected results are summarized in Figure~\ref{fig:accuracy_latency}. As mentioned in Section~\ref{sec:intro}, batching turns out to be inefficient on embedded CPUs due the low number of parallel cores (4 in the Cortex-A53), hence, Seq-TTA is slightly faster than Batch-TTA.
Most importantly, AdapTTA enables substantial acceleration, with much faster prediction rates ranging from 1.40$\times$ to 1.78$\times$ in 5C and from 1.49$\times$ to 2.21$\times$ in 10C.

In MobileNetV1, AdapTTA on 10C outperforms Seq-TTA on 5C in both accuracy (+3.1\% vs. +2.7\%) and speed (4.05\,FPS vs. 3.73\,FPS). The reason can be inferred from Table~\ref{tab:avgruns}, which reports the average number of inferences needed to make a prediction with AdapTTA. AdapTTA needs less than than 5 (4.57) inferences on average (row MobileNetV1, column 10C). Despite that, it achieves superior performance than a static 5C implementation, demonstrating that static TTA is too conservative in most cases and unreliable for less frequent complex inputs.

\begin{table}[t!]
\centering
\caption{Average number of inferences in AdapTTA for the 5-Crops (5C) and 10-Crops (10C) policies.}
\label{tab:avgruns}
\begin{tabular}{lcc}
\toprule
\textbf{ConvNet}                         & \textbf{5C} & \textbf{10C} \\
\midrule
MobileNetV1     & 2.81        & 4.57         \\
MobileNetV2     & 3.37        & 6.26         \\
\bottomrule
\end{tabular}
\vspace{-0.5cm}
\end{table}

\section{Conclusions}\label{sec:conc}
This work presented AdapTTA, an adaptive implementation of TTA suited for embedded systems powered by off-the-shelf CPUs. Differently from static strategies, AdapTTA aims to minimize number of feed-forward passes to output a correct prediction depending on the input complexity. Experimental results proved its efficiency across different ConvNets for the mobile segment and different TTA policies. Overall, AdapTTA reaches substantial acceleration, from to 1.49$\times$ to 2.21$\times$ compared to static TTA policies, with no loss of prediction accuracy.

\bibliographystyle{IEEEtran}
\bibliography{refs}

\end{document}